# Consensus ranking under the exponential model*

**Marina Meilă**
Department of Statistics
University of Washington
Seattle, WA 98195

**Kapil Phadnis**  **Arthur Patterson**
University of Washington
Seattle, WA 98195

**Jeff Bilmes**
Department of Electrical Engineering
University of Washington
Seattle, WA 98195

## Abstract

We analyze the generalized Mallows model, a popular exponential model over rankings. Estimating the central (or consensus) ranking from data is NP-hard. We obtain the following new results: (1) We show that search methods can estimate both the central ranking $\pi_0$ and the model parameters $\theta$ exactly. The search is $n!$ in the worst case, but is tractable when the true distribution is concentrated around its mode; (2) We show that the generalized Mallows model is jointly exponential in $(\pi_0, \theta)$, and introduce the conjugate prior for this model class; (3) The sufficient statistics are the pairwise marginal probabilities that item $i$ is preferred to item $j$. Preliminary experiments confirm the theoretical predictions and compare the new algorithm and existing heuristics.

## 1 Introduction

We are given a set of $N$ *rankings*, or *permutations*[1] on $n$ objects. These rankings might represent individual preferences of a panel of $N$ judges, each presented with the same set of $n$ candidates. Alternatively, they may represent the ranking votes of a population of $N$ voters. The problem of *rank aggregation*, or of finding a *consensus ranking*, is to find a single ranking $\pi_0$ that best "agrees" with all the $N$ rankings. This process can also be seen as a voting rule, where the $N$ voters' preferences are aggregated in an election to produce a consensus order over the candidates, the top ranked being the winner.

* This material is based upon work supported by the National Science Foundation under grant IIS-0535100 and by an ONR MURI grant N000140510388.

[1] We use *permutation* and *ranking* interchangeably.

Various measures of "agreement" have been proposed (for a good overview, see [Critchlow, 1985]). Of these, Kendall's metric [Fligner and Verducci, 1986] has been the measure of choice in many recent applications centered on the analysis of ranked data [Ailon et al., 2005, Cohen et al., 1999, Lebanon and Lafferty, 2002]. The Kendall distance is defined as:

$$d_K(\pi, \pi_0) = \sum_{l \prec_\pi j} 1_{[j \prec_{\pi_0} l]} \quad (1)$$

In the above, $\pi, \pi_0$ represent permutations and $l \prec_\pi j$ ($l \prec_{\pi_0} j$) mean that $l$ precedes $j$ (i.e is preferred to $j$) in permutation $\pi$ ($\pi_0$). Hence $d_K$ is the total number of pairwise disagreements between $\pi$ and $\pi_0$.

This distance was further generalized to a family of parametrized distances $d_\theta(\pi, \pi_0)$ depending on a parameter vector $\theta$ by [Fligner and Verducci, 1986]. Based on these distances, defining probabilistic models of the form $P(\pi) \propto e^{-d_\theta(\pi, \pi_0)}$ is immediate. Estimating $\pi_0$ by e.g Maximum Likelihood (ML) is equivalent to finding the consensus ranking. In fact, various voting rules have been studied in the context of statistical estimation [Conitzer and Sandholm, 2005]. Such estimation problems for generalizations of the Kendall distance are the focus of the present paper.

## 2 Background: Generalized Mallows models

This section follows the excellent paper of [Fligner and Verducci, 1986] which should be consulted for more details. Let $\pi$ denote a permutation over the set $[n] = \{1, 2, 3 \ldots n\}$, where $\pi(l)$ is the rank of element $l$ in $\pi$ and $\pi^{-1}(j)$ is the element at rank $j$. One can uniquely determine any $\pi$ by the $n-1$ integers $V_1(\pi), V_2(\pi), \ldots V_{n-1}(\pi)$ defined as

$$V_j(\pi) = \sum_{l > j} 1_{[l \prec_\pi j]} \quad (2)$$

In other words, $V_j$ is the number of elements in $j+1:n$ which are ranked before $j$ by $\pi$. It follows from the above that $V_j$ takes values in $\{0, \ldots n-j\}$. We note that while the values $\pi(l)$ are dependent, the values $V_j$ may be chosen independently in specifying a $\pi$. In [Feller, 1968] it is moreover shown that the numbers $V_j$ are uniformly distributed if $\pi$ is sampled uniformly.

We say that a distance between permutations $d(\pi, \pi_0)$ is *right-invariant* if $d(\pi\bar{\pi}, \pi_0\bar{\pi}) = d(\pi, \pi_0)$ for any permutation $\bar{\pi}$, where $\pi\bar{\pi}(l) = \pi(\bar{\pi}(l))$. Requiring that a distance is right invariant means that we want it to be indifferent to the relabeling of the $n$ objects, which is a standard assumption. For any right-invariant $d$, we have $d(\pi, \pi_0) = d(\pi\pi_0^{-1}, \mathrm{id}) \equiv D(\pi\pi_0^{-1})$ and therefore the distance is completely determined by the function $D(\pi) \stackrel{def}{=} d(\pi, \mathrm{id})$ where id denotes the identity permutation $\mathrm{id} = (1, 2, \ldots n)$.

### 2.1 Generalized Kendall distance

From (1) and (2) it is easy to see that the Kendall distance has a simple expression $D_K(\pi) = \sum_{j=1}^{n-1} V_j(\pi)$. Therefore, [Fligner and Verducci, 1986] proposed the parametrized generalization of the Kendall distance defined by

$$D_\theta(\pi) = \sum_{j=1}^{n-1} \theta_j V_j(\pi), \quad \theta_j \geq 0 \qquad (3)$$

where $\theta = (\theta_{1:n-1})$ is a parameter vector. The Kendall distance is a metric [Mallows, 1957]. The generalization (3) may be asymmetric unless $\theta_j$ is constant for all $j$. Therefore, $d_\theta$ is not in general a metric.

$D_\theta$ is a versatile and convenient measure of divergence between rankings. By choosing the $\theta$ parameters to e.g decrease with $j$ we can emphasize the greater importance of ranking the first items in $\pi_0$ correctly relative to the correct ranking of items with low ranks in $\pi_0$. Variations of this model where the "emphasized ranks" $j$ can be selected at will are also possible.

### 2.2 Generalized Mallows models

The following family of exponential models based on the divergence (3) is called the *generalized Mallows model* [Fligner and Verducci, 1986]

$$P_{\theta,\pi_0}(\pi) = \frac{e^{-d_\theta(\pi,\pi_0)}}{\psi(\theta)} = \frac{e^{-D_\theta(\pi\pi_0^{-1})}}{\psi(\theta)} \qquad (4)$$

In the above, $\psi(\theta)$ is a normalization constant that does not depend on $\pi_0$. It was shown in [Fligner and Verducci, 1986] that the model (4) factors into a product of independent univariate exponential models, one for each $V_j$ and that

$$\psi(\theta) = \prod_{j=1}^{n-1} \psi_j(\theta_j) = \prod_{j=1}^{n-1} \frac{1 - e^{-(n-j+1)\theta_j}}{1 - e^{-\theta_j}} \qquad (5)$$

$$P[V_j(\pi\pi_0^{-1}) = r] = \frac{e^{-\theta_j r}}{\psi_j(\theta_j)} \qquad (6)$$

The above models are well defined for any real values of the parameters $\theta$. However, we are interested only in the values $\theta_j \geq 0$, for which the probability distribution has a maximum at $V \equiv 0$. This case corresponds to a distribution over orderings where all the high probability instances are small perturbations of the central permutation. For $\theta \equiv 0$, $P_\theta \equiv P_0$ is the uniform distribution. For $\theta_1 = \theta_2 = \ldots = \theta_{n-1}$ (4) is the *Mallows model* [Mallows, 1957]. The size of the $\theta$ parameters controls the concentration of the distribution around its mode $\pi_0$; smaller values make the distribution closer to uniform, while larger values make it more concentrated.

## 3 The ML estimation problem

### 3.1 Parameter estimation.

Assume an independent sample $\pi_{1:N}$ of size $N$ has been obtained from model (4). Then the data log-likelihood can be written as

$$l(\theta, \pi_0) = \ln P(\pi_{1:N}; \theta, \pi_0) \qquad (7)$$

$$= -N \sum_{j=1}^{n-1} \left[ \theta_j \frac{\sum_{i=1}^{N} V_j(\pi_i \pi_0^{-1})}{N} - \ln \psi_j(\theta_j) \right] \qquad (8)$$

$$= -N \sum_{j=1}^{n-1} \left[ \theta_j \bar{V}_j - \ln \psi_j(\theta_j) \right] \qquad (9)$$

In the above $\bar{V}$ is the sample expectation of $V_j(\pi\pi_0^{-1})$. It is easy to see that for any fixed $\pi_0$, the model (4) is an exponential family [DeGroot, 1975] with parameters $\theta$. Moreover, because the random variables $V_j$ are independent, each $V_j$ is distributed according to an exponential model with one parameter $\theta_j$. This is reflected in equation (9) where the log-likelihood $l$ decomposes into a sum of terms, each depending on a single $\theta_j$.

Maximizing the log-likelihood to estimate $\theta$ when $\pi_0$ is known is therefore immediate. It amounts to solving the implicit equation in one variable obtained by taking the partial derivative w.r.t. $\theta_j$ in equation (9). As in [Fligner and Verducci, 1986] this equation is rewritten

$$\bar{V}_j = \frac{1}{e^{\theta_j} - 1} - \frac{n - j + 1}{e^{(n-j+1)\theta_j} - 1}, \quad j = 1 : n - 1 \quad (10)$$

Note that $l(\theta, \pi_0)$ is log-concave in $\theta$. Hence equation (10) has a unique solution for any $j$ and any $\bar{V}_j \in [0, n-j]$ (see e.g [Fligner and Verducci, 1986]). This solution has in general no closed form expression, but can be obtained numerically by standard iterative algorithms for convex/concave optimization [Bertsekas, 1999].

### 3.2 The centroid estimation problem

In the following we study the combinatorial problem of estimating the unknown mode $\pi_0$. Before addressing this, however, we introduce a summary statistic that will prove pivotal to our findings. This is the matrix $Q(\pi_{1:N})$ defined as

$$Q_{jl}(\pi_{1:N}) = \frac{1}{N} \sum_{i=1}^{N} 1_{[j \prec_{\pi_i} l]} \quad \text{for } j, l = 1 : n \quad (11)$$

In other words, $Q_{jl}(\pi_{1:N})$ is the probability that $j$ precedes $l$ in the sample. In the rest of the paper, when no confusion is possible, we will denote $Q(\pi_{1:N})$ simply by $Q$. Also, $Q(\pi)$ denotes the $Q$ matrix corresponding to a single permutation $\pi$. The elements of $Q(\pi)$ are $\{0, 1\}$ valued while the elements of $Q \equiv Q(\pi_{1:N})$ are rational numbers for any finite $N$.

One of the most effective mode estimation procedures is the FV heuristic [Fligner and Verducci, 1988] and can be described in terms of $Q$. Let $\bar{q}_l = \sum_{j=1}^{n} Q_{jl}$. Note that $\bar{q}_l$ is one less than the average rank of $l$ in the data. Let $\bar{\pi}_0$ denote the permutation given by sorting the $\bar{q}_l$ values in increasing order. In [Fligner and Verducci, 1988] it is argued that this permutation is an unbiased estimator of $\pi_0$.

The FV heuristic starts with this permutation, plus the set of all its neighbors at $d_K = 1$; for each of these candidates, the parameters $\theta$ are estimated and the data likelihood computed. The most likely $\pi_0$ of the set is then chosen.

For $\theta_1 = \theta_2 = \ldots = \theta_{n-1} \geq 0$ (the Mallows model) the optimal $\pi_0$ does not depend on $\theta$ and the problem becomes one of finding

$$\pi_0 \in \operatorname*{argmin}_{\pi'} \sum_{i=1}^{N} d_K(\pi_i, \pi') \quad (12)$$

This is precisely the consensus ranking problem. It is known that this problem is NP-hard [Bartholdi et al., 1989], and solving it approximately has been addressed in the literature. The approximation algorithm that guarantees best theoretical bounds is that of [Ailon et al., 2005]; this is a randomized algorithm that achieves a factor 11/7 approximation in minimizing the r.h.s of (12).

In [Cohen et al., 1999] a greedy heuristic (the CSS greedy algorithm) based on graph operations is introduced and tested. The heuristic works under slightly more general conditions, as it assumes that not all of the $n$ items are ranked under all permutations $\pi_i$. A good discussion of the sources of difficulty for this problem is also given. This greedy heuristic achieves a factor 2 approximation. We will return to the CSS heuristic in sections 7 and 8.

Interestingly enough, none of the above tie the concentration of the distribution to the hardness of the problem (recent work that explores the effect of a form of concentration includes [Conitzer et al., 2006, Davenport and Kalagnanam, 2004]). Intuitively, however, the problem should not be difficult if $P_{\theta, \pi_0}$ is concentrated around $\pi_0$. It is also intuitive that if the distribution is uniform, then *any* permutation will be equally qualified to be the mode. The next section exploits exactly this observation.

## 4 Exact ML estimation for $\pi_0$

### 4.1 Estimation of $\pi_0$ for $\theta$ known.

Maximizing the log-likelihood (9) w.r.t $\pi_0$ is the same as minimizing $\sum_{j=1}^{n-1} \theta_j \bar{V}_j$. The following *key observation* allows us to do so. Let us denote for simplicity $\tilde{V}_j(\pi) = V_j(\pi \pi_0^{-1})$. It can be verified that if $\pi_0^{-1}(1) = r$, then $\tilde{V}_1(\pi)$ is the number of elements which come before $r$ in $\pi$. $\bar{V}_1$, the expectation of $\tilde{V}_1$ under the sampling distribution, is the expected number of elements before $r$. Therefore we have:

$$\bar{V}_1 = \sum_{j \neq r} Q_{jr} \quad \text{whenever } \pi_0^{-1}(1) = r \quad (13)$$

Hence, to estimate the first element of $\pi_0$, we can compute all column sums of $Q$ and then choose $\pi_0^{-1}(1) \in \operatorname*{argmin}_{r} \sum_{l \neq r} Q_{lr}$.

This idea can be generalized by induction to all subsequent $j$'s. Assume $\pi_0^{-1} = r_1$ fixed and denote by $\pi|_{-r_1}$ the permutation over $n-1$ elements resulting from $\pi$ if $r_1$ was removed. Again, a simple verification shows that $\tilde{V}_2(\pi)$ represents the number of items that precede $\pi_0^{-1}(2)$ in $\pi|_{-r_1}$. By averaging, it follows that

$$\bar{V}_2 = \sum_{l \neq r_1, r_2} Q_{lr_2} \quad \text{when } \pi_0^{-1}(1:2) = (r_1, r_2). \quad (14)$$

By induction, we obtain

$$\bar{V}_j = \sum_{l \neq r_1, r_2, \ldots r_j} Q_{lr_j} \quad \text{when } \pi_0^{-1}(1:j) = (r_1, r_2, \ldots, r_j).$$
(15)

Therefore, we have in $Q$ the information necessary to find the $\pi_0$ maximizing the likelihood and that an exhaustive search over all the possible permutations can obtain it. One can represent this as a *search tree*, whose nodes represent partial orderings $\rho = (r_1, \ldots r_j)$. Denote by $|\rho|$ the length of the sequence $\rho$. A node $\rho$ with $|\rho| = j$ has $n - j$ children, represented by the sequences $\rho' = \rho|r_{j+1}$ where the symbol | stands for concatenation of sequences and $r_{j+1}$ ranges in $[n] \setminus \rho$ the set complement of $\rho$ in $[n]$. Any path of length $n$ through the search tree starting from the root represents a permutation. A node $(r_1, \ldots r_j)$ at level $j < n$ can be thought of as the set of all permutations that start with $r_1, \ldots, r_j$.

We define the variables of a search algorithm. First,

$$V_j(r_1, r_2, \ldots r_j) = \sum_{l \notin \{r_1, r_2, \ldots r_j\}} Q_{lr_j}. \quad (16)$$

The *cost* at node $\rho = (r_1, \ldots r_j)$ is given by

$$C(r_1, \ldots r_j) = \sum_{l=1}^{j} \theta_l V_l(r_1, \ldots r_l) \quad (17)$$

This cost can be computed recursively on the tree by

$$C(r_1, \ldots r_j) = C(r_1, \ldots r_{j-1}) + \theta_j V_j(r_1, \ldots r_j) \quad (18)$$

The tree nodes can be expanded according to any standard search procedure, such as $A^*$. To direct the search, one also needs a lower bound $A(r_1, \ldots r_j)$ on the *cost to go* from the current partial solution. We will describe possible bounds in the next section. The sum $L(\rho) = C(\rho) + A(\rho)$ represents a lower bound for any permutation in the set prefixed by $\rho$. In such a tree, search can finish with the optimal solution before the whole tree is expanded. Table 1 provides an $A^*$ *Best-First (BF)* search algorithm.

### 4.2 Simultaneous estimation of $\pi_0$ and $\theta$.

Algorithm SEARCHPI can immediately be extended to the more interesting case when both the centroid $\pi_0$ and the parameters $\theta$ are unknown. Recall, for any fixed $\pi_0$ the model (4) is an exponential family model and thus the parameter estimates depend only on the sufficient statistics $\bar{V}_{1:n-1}$. Moreover, the estimate $\theta_j^{ML}$ depends only on $\bar{V}_j$. Hence, any time a node $\rho$ in the search tree is created, $\theta_{|\rho|}^{ML}$ can be readily computed at the node by solving (10) with $\bar{V}_j = V_{|\rho|}(\rho)$.

As mentioned before, this equation does not generally have a closed form solution. However, the values $\theta$ can be tabulated as a function of $\bar{V}$. The value of $\theta_j^{ML}$ in (10) depends only on $\bar{V}_j$ and $n - j$. Therefore, the curve $\bar{V}_{n-j}(\theta)$, and consequently its inverse which we denote $t_{n-j}(\bar{V})$ depend only on $n - j$ and not on $n$. This set of curves, one for each value of $n - j$ can

Table 1: The SEARCHPI algorithm with an admissible heuristic $A$. Node $\rho$ stores: $\rho = r_1, \ldots, r_j$, $j = |\rho|$, $V_j(\rho)$, $\theta_j$, $C(\rho)$, $L(\rho)$; $S$ is the priority queue holding the nodes to be expanded.

**Algorithm** SEARCHPI

**Initialize**

$S = \{\rho_\emptyset\}$, $\rho_\emptyset$ = the empty sequence, $j = 0$, $C(\rho_\emptyset) = V(\rho_\emptyset) = L(\rho_\emptyset) = 0$

**Repeat**

remove $\rho \in \underset{\rho \in S}{\operatorname{argmin}} L(\rho)$ from $S$

if $|\rho| = n$ *(Return)*

    **Output** $\rho$, $L(\rho) = C(\rho)$ and **Stop**.

else *(Expand $\rho$)*

    for $r_{j+1} \in [n] \setminus \rho$
        create node $\rho' = \rho|r_{j+1}$
        $V_{j+1}(\rho') = V_j(r_{1:j-1}, r_{j+1}) - Q_{r_j r_{j+1}}$
    compute $V^{min} = \underset{r_{j+1} \in [n] \setminus \rho}{\min} V_{j+1}(\rho|r_{j+1})$

    calculate $A(\rho)$
    for $r_{j+1} \in [n] \setminus \rho$

        $\theta_{j+1} = t_{n-j-1}(V_{j+1}(\rho'))$
        $C(\rho') = C(\rho) + \theta_{j+1} V_{j+1}(\rho')$
        $L(\rho') = C(\rho') + A(\rho)$
        store node $(\rho', j+1, V_{j+1}, \theta_{j+1}, C(\rho'), L(\rho'))$ in $S$

be computed off-line once and then used for any data with $n$ up to a preset maximum value.

## 5 Computational aspects

### 5.1 Admissible heuristics

We now describe possible functions $A(\rho)$ to be used in place of the cost to go. Such a function needs to satisfy two conditions: to be easily computable, and to lower bound the true cost to go. The simplest heuristic is evidently $A(\rho) = 0$.

**Admissible heuristic for $V$ with known $\theta$.** If the parameters $\theta$ are known, then we only need to find lower bounds on $V_{j'}$ for $j' > j$. When node $\rho$ is expanded, after computing $V_{j+1}$ for all children, we find the minimum over these values as

$$V^{min} = \min_{r \in [n] \setminus \rho} V_{j+1}(\rho|r). \quad (19)$$

For $j + 1 < j' < n - 1$, the best $V_{j'}$ on the current branch are column sums of sub-matrices of $Q$. Letting $(r_{j+1}, r_{j+2}, \ldots, r_{j'})$ be any length $j' - j$ continuation,

we get:

$$V_{j'}(\rho|(r_{j+1},\ldots,r_{j'}))$$
$$= \sum_{i\in[n]\setminus\rho} Q_{ir_{j'}} - \sum_{i\in\{r_{j+1}\ldots r_{j'}\}} Q_{ir_{j'}}$$
$$\geq \max[V^{min} - (j'-j)Q^{max}, 0] = a_{j'}(\rho)$$

where $Q^{max} = \max_{jl} Q_{jl}$ is computed off line. Then $A$ can be computed as $A(\rho) = \sum_{j'=j+1}^{n-1} \theta_{j'} a_{j'}(\rho)$.

**Admissible heuristic for $V$ with constant $\theta$.** For the special case of consensus ranking, when $\theta_j \equiv 1$, an even better heuristic can be used. Sort the off-diagonal values of $Q_{lr}$ in increasing order, denoting the resulting sequence by

$$q_{(1)} \leq q_{(2)} \leq \ldots \leq q_{(n(n-1)/2)} \quad (20)$$

The cost to go in consensus ranking is independent of $\theta$ and equal to $V_{j+1}(\rho'_{j+1}) + V_{j+2}(\rho'_{j+2}) + \ldots V_{n-1}(\rho'_{n-1})$ on some (unknown) path from the current node to the bottom of the tree. Since each $V_j$ is the sum of $n-j$ off-diagonal $Q_{lr}$'s, this cost to go is equal to the sum of $(n-j-1)(n-j-1+1)/2$ distinct off-diagonal elements of $Q$. Hence $A(\rho) = \sum_{l=1}^{(n-j-1)(n-j)/2} q_{(l)}$ is always lower bounding the cost to go. This heuristic depends only on the level $j$ and can be entirely computed before the search.

**Admissible heuristics for unknown $\theta$.** If the parameters $\theta_j$ are estimated simultaneously with the central permutation $\pi_0$, then lower bounding the cost to go requires us to find lower bounds on the parameters $\theta_{j'}$, with $j' > j = |\rho|$ the current level.

Any non-zero lower bound on $\theta_{j'}$ can then be combined with the lower bounds on $V_{j'}$ described above produce an admissible $A$. The derivation of possible lower bounds for the parameters is in [Meilă et al., 2007]. In this case too, the bounds will be computed off-line and will depend only on the tree level $j$.

### 5.2 Number of node expansions

Let us further analyze the algorithm SEARCHPI from a computational point of view. BF algorithms with admissible heuristics are guaranteed to find the optimal solution given enough time. The stopping condition is met when the most promising node is a terminal node. This condition can be met before all nodes in the search tree are expanded. Hence, an important performance parameter for a BF algorithm is the number of nodes that it visits before it finds the optimum. This number clearly depends on the quality of the heuristic – the better a lower bound is $A$ on the cost to go, the more nodes can be pruned from the search tree.

In our case, the worst case running time will be $n!$. The lower limit on the number of nodes created, given by the path of the greedy search strategy, is $n + (n-1) + \ldots + 2 = n(n+1)/2 - 1$. The number of nodes expanded by the greedy strategy is one node in each level, i.e a total of $n-1$ nodes.

A qualitative examination of the cost (17) shows that the larger the value of $\theta_j$, the greater the advantage of the best $r_j$ w.r.t the second best. Hence, large values of $\theta_j$ imply that the chance of a non-optimal subtree at level $j$ to contain the optimal solution is small. In other words, when the values of the parameters are large, which corresponds to a distribution $P_\theta$ that decays fast away from the mode $\pi_0$, then the number of nodes explored will be small. For any admissible heuristic $A$, there are parameters $\theta^{ML}$ for which the BF algorithm will explore exactly the same nodes as the greedy algorithm and no more.

At the other end of the spectrum, if $\theta_j \approx 0$ for all $j$, the search is likely to be intractable. In this case are data sets sampled from an almost uniform distribution, which will have all values $Q_{lr} \approx 0.5$. Data sampled from multi-modal distributions can also fall under this category[2]. For multi-modal distributions, individual $Q_{lr}$ values can take extreme values near 0 or 1, but because no consensus exists, the average $Q_{lr}$ along a column or sub-column will be near 0.5 as well.

In this latter case, the algorithm can be stopped any time, and it will provide the best solution it was able to find so far. For this case, practical optimization usually involves inadmissible heuristics (e.g. beam search). We leave this avenue open for further research.

### 5.3 Number of operations per node.

Upon creating node $\rho' = \rho|r_{j+1} = r_1, \ldots, r_{j+1}$, the SEARCHPI algorithm needs to compute the value of $\bar{V}_{\rho'} = \sum_{l\in[n]\setminus\rho} Q_{lr_{j+1}}$. Computing this sum explicitly takes $\mathcal{O}(n-j)$ operations, which makes the time of exploring one vertical path to the terminal of a tree be $\mathcal{O}(n(n-1) + (n-1)(n-2) + \ldots + 2) = \mathcal{O}(n^3)$. However, by better organizing the data we can obtain a *constant* computation time per node.

$$V_{j+1}(r_1,\ldots,r_{j+1}) = \sum_{l\neq r_{1:j-1},r_{j+1}} Q_{lr_{j+1}} - Q_{r_j r_{j+1}}$$
$$= V_j(r_1,\ldots,r_{j-1},r_{j+1}) - Q_{r_j r_{j+1}} \quad (21)$$

The node $(r_1,\ldots,r_{j-1},r_{j+1})$ is a sibling of $(r_1,\ldots,r_{j+1})$'s parent (hence an "uncle"). In our algorithm, and in any search algorithm that creates all children of a node at once, this node will have been created and its $V$ value available by the time we need to compute $V_{j+1}(r_1, \ldots, r_{j+1})$.

---
[2] In this case, since there is no true parameter $\theta$, we refer to the estimated $\theta^{ML}$

To use this value, we must only make sure that no nodes are deleted from memory while their $V$ values are still needed. This can be achieved with a counter variable associated with each $V_j(\rho)$ which signals when the value is no longer needed. Another possible solution is to pass the $V$ values alone, as tables, down the tree. This way any node can be deleted independently of the rest of the tree. Keeping a table at a node adds a storage of $n - j$ per node.

Selecting the next node can be done efficiently if all the nodes are kept in a priority queue sorted by $L(\rho)$. Fibonacci heaps can attain constant access time, while our STL based implementation uses a binomial heap with access time logarithmic in the length of the queue.

## 6 Identifiability and conjugate prior

### 6.1 Identifiability

The matrix $Q$ represents the sufficient statistics for the parameters $\pi_0, \theta$. Because by definition

$$Q_{lj} + Q_{jl} = 1 \text{ for } l \neq j, \ Q_{jj} = 0 \quad (22)$$

the number r of free parameters in $Q$ is at most $n(n-1)/2$.

The set $\mathcal{Q} = \{Q\}$ of matrices satisfying (22) is a convex polytope, with $n!$ extreme points given by $Q(\pi) = [1_{[l \prec_\pi j]}]_{lj}$. By the Caratheodory theorem [Rockafellar, 1970], any $Q$ in the polytope can be represented by a convex combination of at most $n(n-1)/2+1$ extreme points. This implies that $Q$ can be approximated arbitrarily closely by finite data sets with $N$ large enough. So, asymptotically, any $Q \in \mathcal{Q}$ can represent a set of sufficient statistics.

Note also that for any $Q \in \mathcal{Q}$ and for any permutation $\pi_0$, there is a unique parameter vector $\theta^{ML}(\pi_0) \in \underset{\theta}{\operatorname{argmax}} P_{\theta,\pi_0}(Q)$ (because equation (10) has a unique solution). The following result says that for any data set there is a non-negative $\theta$ estimate.

**Proposition 1** *For any $Q \in \mathcal{Q}$ there exists a permutation $\pi_0$, so that $\theta_j^{ML}(\pi_0) \geq 0$.*

**Proof.** Since $Q_{jl} = 1 - Q_{lj}$ we have $\sum_{jl} Q_{jl} = n(n-1)/2$; therefore there is at least one column $r$ for which $\sum_l Q_{lr} \leq (n-1)/2$. For this column, equation (10) with $j = 1$ will have a non-negative solution $\theta_1$. We now delete column and row $r$ from $Q$ and proceed recursively for $j = 2 : n - 1$. □

This proposition justifies our focusing on the domain of non-negative $\theta$. It shows that such a restriction is not only convenient, it is also necessary to ensure that the model is identifiable. A model $P_{\theta,\pi_0}$ with $\theta > 0$ is *strongly unimodal*; in such a model the probability of any inversion w.r.t $\pi_0$ is less than 0.5 [Fligner and Verducci, 1988].

While almost[3] each $Q \in \mathcal{Q}$ defines uniquely a pair $\theta^{ML}, \pi_0^{ML}$, the converse is not true. There are an infinity of matrices $Q$ which produce the same $\theta^{ML}, \pi_0^{ML}$.

### 6.2 The conjugate prior

The existence of finite sufficient statistics implies that $P_{\theta,\pi_0}(\pi)$ is an exponential family model jointly in $(\theta, \pi_0)$. As such, it will have a conjugate prior, whose form is given below.

**Proposition 2** *Let $\Gamma \in \mathcal{Q}, \nu > 0$; denote $\Gamma_\infty = Q(\mathrm{id}) \in \mathcal{Q}$, $\Theta = \mathrm{diag}(\theta, 0) \in \mathbb{R}^{n \times n}$ and $\Pi_0$ the permutation matrix associated to permutation $\pi_0$. Then*

$$P(\pi_0, \theta\,;\,\nu, \Gamma) \propto e^{-\nu[\mathrm{trace}\,\Gamma_\infty \Pi_0 \Gamma \Pi_0^T \Theta + \ln \psi(\theta)]} \quad (23)$$

*is a conjugate prior for the parameters $(\theta, \pi_0)$ of model (4).*

**Proof.** $V_j(\pi \pi_0^{-1})$ can be written as element $(j,j)$ of $\Gamma_\infty \Pi_0 Q(\pi) \Pi_0^T$ and consequently $\ln P_{\theta,\pi_0}(\pi) = \mathrm{trace}\,\Gamma_\infty \Pi_0 Q(\pi) \Pi_0^T \Theta + \ln \psi(\theta)$. Moreover, $NQ = \sum_{i=1}^N Q(\pi_i)$. Hence,

$$\begin{aligned} P(\pi_0, \theta\,|\,\pi_{1:N}) &\propto P(\pi_{1:N}|\pi_0, \theta) P(\pi_0, \theta\,;\,\nu, \Gamma) \\ &\propto e^{-(N+\nu)[\mathrm{trace}\,\Gamma_\infty \Pi_0 \frac{NQ+\nu\Gamma}{N+\nu} \Pi_0^T \Theta + \ln \psi(\theta)]} \\ &= P(\pi_0, \theta;\,N+\nu, \frac{NQ+\nu\Gamma}{N+\nu}) \quad (24) \end{aligned}$$

We have shown that the distribution in (23) is closed under sampling, in other words it is a conjugate prior [DeGroot, 1975]. It remains to show that the prior is integrable on $\theta_j \geq 0, j = 1 : n - 1$. This is straightforward and left to the reader. □

We note that the general form of a conjugate prior family is $P(\pi_0, \theta\,;\,\nu, \Gamma) \propto h(\theta, \pi_0) e^{-\nu[\mathrm{trace}\,\Gamma_\infty \Pi_0 \Gamma \Pi_0^T \Theta + \ln \psi(\theta)]}a$ where $h(\theta, \pi_0)$ is a function that renders the prior integrable and doesn't depend on $\nu, \Gamma$. Our proposition extends immediately to this case as well.

The prior above is defined up to a normalization constant. At present we do not have a closed form formula for this constant. We also stress that the sufficient statistics $Q$ for the model (4) *are not minimal* and the model itself, in the above parametrization, is not a minimal exponential model.

---
[3]Except for those $Q$ for which there are ties in $\pi_0$.

It is interesting, nevertheless, to interpret the prior's parameters. The $\nu$ parameter's role as "equivalent sample size" is obvious; let us now look at the matrix parameter $\Gamma$. If $\Gamma$ represents an expectation matrix $E_{\theta^*,\pi^*}[Q(\pi)]$ under model (4) the conjugate prior is equivalent to having seen $\nu$ samples from a distribution centered at $\pi^*$ with spread $\theta^*$. If one uses another $\Gamma$ in the prior, that corresponds to having seen $\nu$ samples from a distribution not in the class represented by (4).

The matrix $\Gamma_0$ obtained from $\theta^* \equiv 0$ has $(\Gamma_0)_{ij} = 0.5$ in each off-diagonal entry. This matrix corresponds to an non-informative prior w.r.t $\pi_0$, as $\theta^* \equiv 0$ represents the uniform distribution. Using a conjugate prior with $\Gamma_0$ implements a smoothing over the parameters while being non-informative w.r.t the central permutation. It can be easily verified that any other $\Gamma \in \mathcal{Q}$ is informative w.r.t both $\theta$ and $\pi_0$. Hence, in the conjugate prior framework it is impossible to express ignorance w.r.t to the central distribution, while expressing knowledge about the parameters $\theta$.

From an algorithmic standpoint, working with the conjugate prior is, as expected, straightforward. The full posterior, up to the normalization constant, is obtained as a summation of sufficient statistics and prior parameters. This allows one to compare the posteriors of any two models. If one is interested in the Maximum A-Posteriori (MAP) estimate, this can be readily obtained by algorithm SEARCHPI with $Q$ replaced by $(NQ + \nu\Gamma)/(N + \nu)$.

## 7 Experiments

The experiments in this section evaluate various existing algorithms on the consensus ranking problem of estimating $\pi_0$. Since estimating $\theta$ adds only a small constant time per search step, we consider that this case embodies the core difficulties of the estimation problem. Exception would make the cases when $\theta_j$ has comparatively large values at large $j$'s, signifying that the most important stages of the ranking are among the *last* ones, while getting the highly ranked elements of $\pi_0$ is less important. This case is rather unrealistic in practice.

We implemented the SEARCHPI in C++ with the heuristics mentioned in section 5. This algorithm is denoted in the experiments as BF. We also implemented a sub-optimal search algorithm that runs the SEARCHPI for a predefined amount of time (5 minutes) then continues with greedy search from the largest level $j$ attained in the BF search. This algorithm is denoted BF-CSS (the greedy search is denoted by CSS as described below).

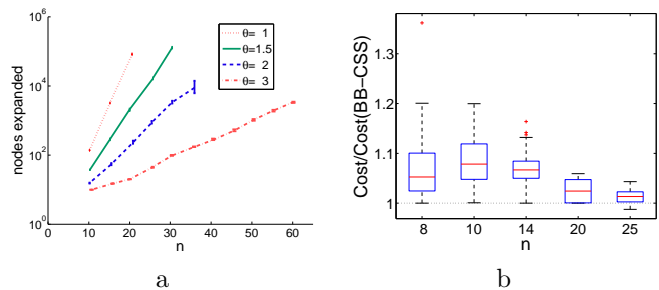

Figure 1: (a) The average number of nodes expanded by the SEARCHPI with heuristic $A = 0$ for various values of $n$ and $\theta$. The error bars mark the minimum and maximum values over $n_{iter} = 10$ replications. (b) The cost of the greedy CSS heuristics as a fraction of the BF-CSS cost. The BF-CSS heuristic is in effect the exact BF algorithm for $n \leq 14$. The data are $Q$ matrices with independent random entries. The boxplots are over $n_{iter} = 10$ replications.

Although theoretically the search time should not depend on the true $\pi_0$ in all our experiments we select a random $\pi_0$ every time in order to average out any artifacts of the implementation (for example, having the first branch always be the optimal one could make the algorithm faster). We also mention that our implementation of the SEARCHPI is a pilot implementation not optimized w.r.t running time.

The other algorithms we compared were the FV heuristic of [Fligner and Verducci, 1988], the GREEDY-ORDER algorithm of [Cohen et al., 1999] (here denoted CSS) and the algorithm of [Ailon et al., 2005] (denoted ACN here). Our implementation of the FV heuristic omits the search around $\bar{\pi}$ and therefore has a run-time complexity of $\mathcal{O}(n^2)$. The ACN algorithm is also $\mathcal{O}(n^2)$ while the greedy algorithm is $\mathcal{O}(n^3)$. In our experiments, these algorithms ran very fast (fractions of a second) in all the experiments performed.

**Experiments with concentrated distributions** As mentioned in section 2, the consensus ranking problem has two regimes. In the *asymptotic* regime the distribution is concentrated around its mode ($\theta^{ML}$ is large), and $N$ is large enough that $\pi^{ML}$ coincides with the true $\pi_0$. This is an easy case for the BF search, but it is also an easy case for all heuristic algorithms mentioned in section 2.

We have confirmed this experimentally, on samples with $N = 5000$ from distribution $P_{\theta,\pi_0}$ with random $\pi_0$ and with $\theta \equiv 1, 1.5, 2, 3$. Each experiment was replicated $n_{iter} = 10$ times. In all cases, all the heuristics returned the optimal permutation. For this experiment, Figure 1,a shows the number of nodes expanded by the BF algorithm as a function of $\theta$ and $n$.

We also ran a comparison of the heuristics FV, ACN, CSS on samples of size $N = 5000$ from a distribution with $\theta = 0.3$ (only moderately concentrated) and with $n = 10, \ldots 50$. Each experiment was replicated $n_{iter} = 100$ times. For up to $n = 40$, all the heuristics returned the true permutation $\pi_0$. For these experiments, the optimal permutation was not known except for $n \leq 15$ but the large $N$ ensures that with high probability the optimum coincides with the true $\pi_0$.

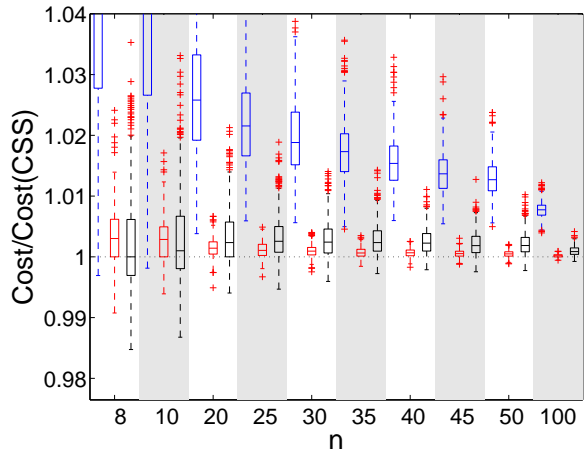

Figure 2: The cost (from left to right) of the true $\pi_0$, the FV and the ACN heuristics, as fractions of the CSS cost. The data are $N = 100$ permutations from $P_{0.03,\pi_0}$ with random $\pi_0$. The boxplots are over $n_{iter} = 500$ replications.

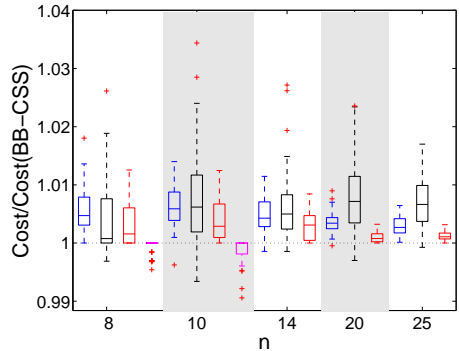

Figure 3: The cost (from left to right) of the FV, ACN, CSS, and SEARCHPI (only for $n = 8, 10$) algorithms, as fractions of the BF-CSS cost. The data are as in Figure 2. The boxplots are over $n_{iter} = 50$ replications.

**Experiments with almost uniform distributions** At the other end of the spectrum is the *combinatorial* regime, where the observed permutations are distributed almost uniformly ($\theta \approx 0$) and $N$ is relatively small so that the true $\pi_0$ is different from $\pi_0^{ML}$. We have simulated this case by generating $N = 100$ samples from a model with $\theta = 0.003$. The distribution being practically indistinguishable from uniform, and the $Q_{ij}$ values being very close to 0.5, the differences in cost between various solutions are minute, and they are presented only as surrogates of a quality of the search, since the optimal $\pi_0$ is not known. For the same reason, all algorithms except SEARCHPI have been compared on $n_{iter} = 500$ replicated experiments.

The comparison between the heuristic algorithms is presented in Figure 2. The greedy CSS heuristic is consistently the best at all scales. Its advantage over the randomized algorithm of ACN is increasing with larger $n$. The true model $\pi_0$ is never optimal for this data distribution, while its estimate by the FV heuristic is better but loses to the other algorithms. The "shrinking towards" 1 effect observed for larger $n$ reflects the fact that a the larger number of values in $Q$ are near 0.5 when $n$ is large. This in turn shrinks the range between the maximum and minimum cost.

Figure 3 shows comparisons between the three heuristics, the optimal BF (for $n = 8, 10$ only) and the approximate search BF-CSS. The costs are plotted as fractions of the cost BF-CSS. Therefore, the optimal BF cost always appears below or equal to 1. The experiments also show that in a large number of cases, the suboptimal BF-CSS outperforms all the other algorithms and improves on the closely related CSS greedy algorithm.

We do not claim the BF-CSS to be the ultimate approximate search heuristic. Better and faster suboptimal searches (e.g beam-search) could be implemented. We only demonstrate by BF-CSS that the search tree approach is effective in improving the cost, or alternatively, in getting closer to a consensus, over the traditional heuristics.

**Experiments with no consensus and large range of $Q$.** In this set of experiments, the data consists of a matrix $Q$ with elements randomly sampled from $[0, 1]$ subject to the constraint $Q_{ij} + Q_{ji} = 1$ and 0 diagonal. This simulates the case of a multi-modal distribution, where the permutations exhibit no consensus, but are also non-uniform. Such a setting was examined experimentally by [Cohen et al., 1999]. In this problem, because the cost $C$ can vary significantly with the choice of $\pi_0$, finding a central permutation $\pi_0$ minimizing this cost is a legitimate practical question. For instance, this task is a subtask of learning to rank in [Cohen et al., 1999].

The experimental setting is identical to the previous, except that the experiments are now replicated $n_{iter} = 10$ times. Figure 1,b shows the costs, as a fraction of the cost of BF-CSS. Similarly to 3, the BF algorithm improves on all heuristics for small $n$ and the suboptimal BF-CSS improves by a few percent over the greedy algorithm (the best contender of the other heuristics) for larger values of $n$.

In the interest of fairness, we stress once more that the FV algorithm could be improved by local search like in [Fligner and Verducci, 1988] and that the CSS algorithm can also be improved by first finding the *strongly connected components* as described in [Cohen et al., 1999].

## 8 Related work and Discussion

This work builds on [Fligner and Verducci, 1986] and [Fligner and Verducci, 1990] who introduced the generalized Mallows model and exploited the fact that it is an exponential family model in $\theta$ alone. As such, they use a conjugate prior on $\theta$ with a uniform prior on $\pi_0$. We have shown in section 6 such a prior is *not* the conjugate prior for $\theta, \pi_0$ jointly. The normalization constant for their posterior is not computable in closed form, and it has strong similarities with the normalization constant of (23), suggesting that the latter may not be computable in closed for either. Another notable spin-off of [Fligner and Verducci, 1990] is [Lebanon and Lafferty, 2002] where the posterior of [Fligner and Verducci, 1990] is used as a conditional probability model over permutations, to be estimated from data by a MCMC algorithm. Other exhaustive procedures for computing consensus rankings have been developed as well. In [Davenport and Kalagnanam, 2004], a greedy heuristic and branch-and-bound procedure is developed for computing the consensus ranking based on the pairwise winner-looser graph. This procedure was extended in [Conitzer et al., 2006], which utilizes not only graphs but also linear programming approximations leading to better bounds. These papers empirically explore the effect of concentration based on a single probability of a deviation from pairwise preferences in $\pi_0$. They also find that as concentration increases, compute-time decreases.

We have presented a new algorithm and a comparison of algorithms from various fields on the estimation of the consensus ranking. Our approach to concentration is based on the parameters of an exponential model. While our algorithm is certainly optimal, it is also by far the slowest. Experiments have highlighted the existing trade-offs: in the asymptotic regime, all heuristics work well; using SEARCHPI is also efficient. In the combinatorial case, if we are interested in the cost only, then the differences in cost are so minute that almost any heuristic (even no optimization) will be acceptable. In other words, while the problem of consensus ranking is theoretically NP hard, minimizing the cost (approximately) is practically easy.

What is hard is finding the individual permutation that achieves best consensus in the combinatorial regime. If this is of interest, then our experiments have shown that the existing heuristics differ and that the SEARCHPI outperforms the other contenders when it's tractable. We are currently implementing faster and non-admissible versions of SEARCHPI, with the expectation that, even if exact optimization is not tractable, using a search like in SEARCHPI for a pre-specified time can improve over greedy search.

We can show (proof omitted) that the GREEDY-ORDER algorithm of [Cohen et al., 1999] is the greedy counterpart of the SEARCHPI algorithm. In this sense, the good results of the CSS heuristic for larger $n$ suggest that adding an amount of search to this already good heuristic is worthwhile.

We conclude by pointing out that with real ranking data we expect to encounter few unimodal distributions. We plan to continue this work toward the more ambitious goal of estimating parametric and non-parametric mixtures over the space of rankings.